%% file: root.tex
\documentclass[letterpaper, 10 pt, conference]{ieeeconf}  % Comment this line out if you need a4paper

\IEEEoverridecommandlockouts                              % This command is only needed if 
\usepackage{graphics} % for pdf, bitmapped graphics files
\usepackage{epsfig} % for postscript graphics files
\usepackage{mathptmx} % assumes new font selection scheme installed
\usepackage{times} % assumes new font selection scheme installed
\usepackage{amsmath} % assumes amsmath package installed
\usepackage{amssymb}  % assumes amsmath package installed
\usepackage{booktabs}
\usepackage{xcolor}

\usepackage{cleveref}

\usepackage[ruled,vlined]{algorithm2e}

\usepackage{titlesec}
\usepackage{microtype}
\usepackage{ragged2e}

\usepackage[ruled,vlined]{algorithm2e}

\SetKwSty{textsf}

\renewcommand{\FuncSty}[1]{\texttt{#1}}

\SetCommentSty{\color[gray]{0.8}\textit}

\title{\LARGE \bf
Reflection-Based Task Adaptation for Self-Improving VLA
}

\author{
    Baicheng Li*$^{1,2}$ \quad
    Dong Wu$*^{1}$ \quad
    Zike Yan$^{3}$ \quad
    Xinchen Liu$^{2}$ \quad
    Lusong Li*$^{2}$ \quad
    Zecui Zeng*$^{2}$ \quad
    Hongbin Zha*$^{1}$
    \\[1ex]
    $^{1}$School of Intelligence Science and Technology, Peking University \\
    $^{2}$JD Explore Academy \quad $^{3}$AIR, Tsinghua University
    \\[1ex]
}

\begin{document}

\maketitle

\thispagestyle{empty}
\pagestyle{empty}

\input{sec/abstract}
\input{sec/intro}

\input{sec/related_work}

\input{sec/method}

\input{sec/experiments}

\section{CONCLUSIONS}

In this paper, we introduced Reflective Self-Adaptation, a novel framework that addresses the challenge of in-situ adaptation for pre-trained VLA models. Our dual-pathway architecture enables an agent to autonomously improve by learning from its own failures and successes. The failure-driven pathway leverages a VLM’s causal reasoning to dynamically synthesize dense rewards from failures, accelerating RL exploration. This is complemented by a success-driven SFT pathway that performs quality-guided imitation of the best successes to ensure stability and prevent reward hacking. Our experiments on challenging manipulation tasks demonstrated substantially faster convergence and higher final success rates than representative baselines. This work presents a robust solution for creating self-improving agents, and we believe that endowing robots with reflection is a promising direction for truly adaptive machines.

\clearpage

\bibliographystyle{IEEEtran}
\bibliography{my_references}

\end{document}

%% file: sec/abstract.tex
\begin{abstract}

Pre-trained Vision-Language-Action (VLA) models represent a major leap towards general-purpose robots, yet efficiently adapting them to novel, specific tasks in-situ remains a significant hurdle. While reinforcement learning (RL) is a promising avenue for such adaptation, the process often suffers from low efficiency, which hinders the rapid mastery of the task. In this paper, we introduce Reflective Self-Adaptation, a novel framework designed to achieve rapid, autonomous task adaptation without human intervention. Our framework establishes a self-improving loop where the agent systematically learns from its own experience to enhance both its strategy and execution.

The core of our framework is a dual-pathway architecture that addresses the full adaptation lifecycle. First, a Failure-Driven Reflective RL pathway enables rapid learning by using the VLM's causal reasoning to automatically synthesize a targeted, dense reward function from failure analysis. This provides a focused learning signal that significantly accelerates policy exploration. However, optimizing such proxy rewards introduces a potential risk of "reward hacking," where the agent masters the reward function but fails the actual task. To counteract this, our second pathway, Success-Driven Quality-Guided SFT, grounds the policy in holistic success. It identifies and selectively imitates high-quality successful trajectories, ensuring the agent remains aligned with the ultimate task goal. This pathway is further strengthened by a conditional curriculum mechanism to overcome initial exploration challenges.

We conduct experiments in a series of challenging manipulation tasks. The results demonstrate that our framework achieves substantially faster convergence and higher final success rates compared to representative baselines. Our work presents a robust solution for creating self-improving agents that can efficiently and reliably adapt to the complexities of new environments.

\end{abstract}

%% file: sec/intro.tex
\section{Introduction}

The advent of large-scale foundation models has catalyzed a paradigm shift in robotics, giving rise to Vision-Language-Action (VLA) models capable of performing a wide array of manipulation tasks from natural language instructions. Pre-trained on vast datasets of internet-scale robot trajectories, such models have demonstrated remarkable zero-shot generalization, hinting at a future of general-purpose robots. However, a significant gap persists between this generalist capability and reliable execution in novel or specific environments. When a pre-trained VLA is deployed in a new setting, such as a user's home, its initial success rate on a given task is often unacceptably low. This "last-mile" adaptation problem is a critical barrier to practical deployment. While reinforcement learning (RL) is a promising avenue for such in-situ improvement, the approach is often hampered by significant sample inefficiency, which makes achieving task proficiency a protracted and challenging process.

To bridge this gap, we ask: Can an agent learn to systematically improve its performance in a new environment, by reflecting on its own experiences without recourse to human experts? In this work, we provide a strong affirmative answer by proposing the Reflective Self-Adaptation framework, a self-improving closed loop that learns synergistically from both its failures and successes. The core philosophy of our framework is to treat failures not just as setbacks, but as invaluable learning opportunities for understanding why a strategy is flawed, and to treat successes not as mere goals, but as templates of how a task can be accomplished with excellence. Our framework operationalizes this philosophy through a dual-pathway architecture. The Failure-Driven Reflective RL pathway utilizes the VLM to analyze failures and automatically synthesize a dense reward, which provides targeted guidance to significantly accelerate learning. Complementing this, the Success-Driven Quality-Guided SFT pathway addresses the potential risk of "reward hacking" by grounding the policy in high-quality successes. This imitation-based refinement also constrains the exploration space for RL, steering it towards promising solutions and further boosting convergence speed.

The main contributions of our work are threefold:
\begin{itemize}
\item We propose the Reflective Self-Adaptation framework, a novel dual-pathway architecture for autonomous, in-situ VLA adaptation. It provides a complete solution that systematically learns from both failures and successes to achieve rapid and robust task mastery without human intervention.
\item We introduce a Failure-Driven Reflective RL pathway, centered around our novel Reflective Reward Synthesis method. This method leverages the VLM's causal reasoning to automatically generate dense rewards from failure analysis, transforming intractable reward engineering into a structured, automated reasoning task.
\item We design a complementary Success-Driven Quality-Guided SFT pathway. This mechanism stabilizes learning and ensures goal alignment by selectively imitating high-quality successes, robustly mitigating the risks of reward hacking and cold-start exploration through intrinsic quality assessment and a VLM-driven curriculum.
\end{itemize}

Through extensive experiments, we show that our framework enables a VLA agent to autonomously master complex tasks, dramatically improving performance over representative baselines. Our findings suggest that endowing agents with the capacity for reflection is a promising direction for building the next generation of adaptive and intelligent robots.

\begin{figure*}[ht]
    \centering
    \includegraphics[width=\textwidth]{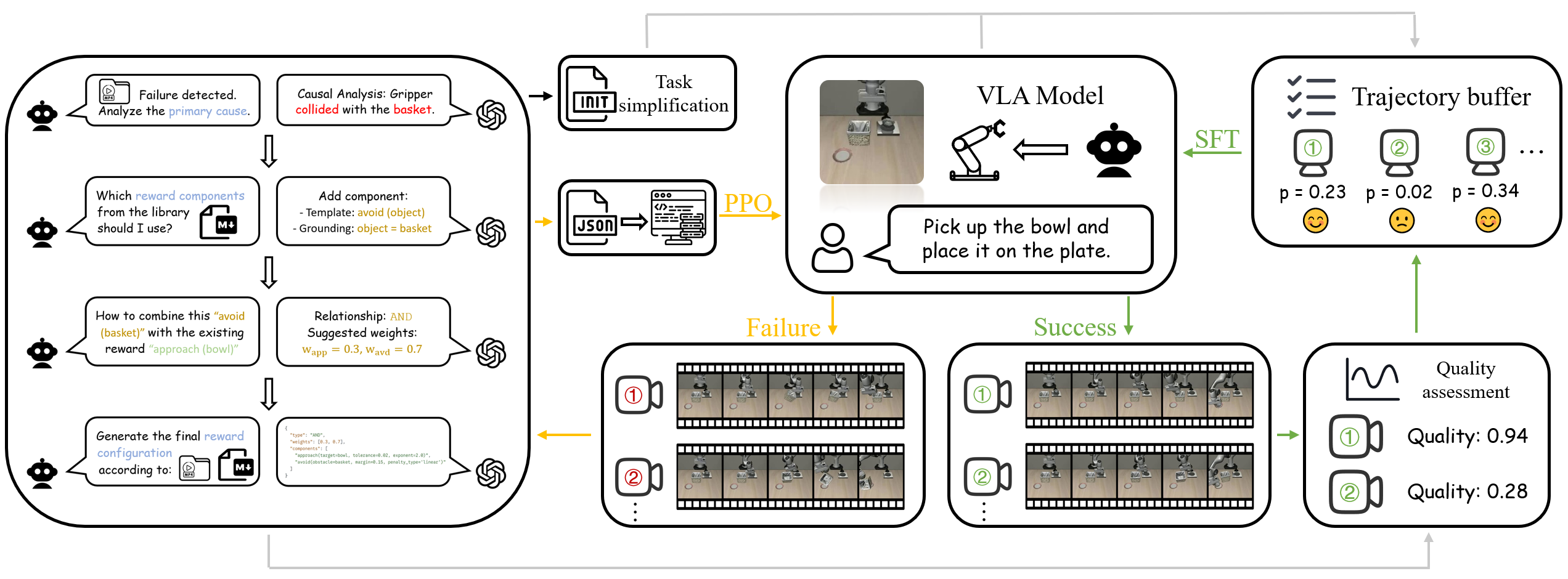}
    \caption{Our Reflective Self-Adaptation framework. Failure loop (yellow): A VLM reflection dialogue synthesizes dense rewards for PPO updates. Success loop (green): Performs prioritized SFT guided by an intrinsic quality assessment.}
    \label{fig:pipeline}
\end{figure*}

%% file: sec/related_work.tex
\section{Related Work}
\label{sec:related_work}
\subsection{Vision-Language-Action (VLA) Models}
Vision-Language-Action (VLA) models aim to create general-purpose robots by framing manipulation as a sequence modeling problem. Architecturally, these policies are either trained as high-capacity encoder-decoder Transformers on large-scale robotics data~\cite{brohan2022rt, team2024octo, reuss2024multimodal, jiang2023vima}, or more recently, built by leveraging powerful web-scale Vision-Language Models (VLMs) as a commonsense reasoning backbone~\cite{zitkovich2023rt, kim2024openvla, driess2023palm, zhao2025cot, belkhale2024rt, black2024pi0}. To generate motor commands, these models either discretize the action space into a vocabulary of predictable tokens~\cite{brohan2022rt, zitkovich2023rt, kim2024openvla} or generate continuous actions, where diffusion models have become a state-of-the-art method for producing complex trajectories~\cite{chi2023diffusion, team2024octo, huang2025vlmtdp, bjorck2025gr00t}. While demonstrating impressive zero-shot generalization, the knowledge of these VLAs is acquired from static, offline datasets and remains fixed after pre-training, leaving them unable to adapt in-situ. Our work directly addresses this critical gap, introducing a framework that enables VLAs to learn and self-improve from their own ongoing experience in new environments.

\subsection{Reinforcement Learning with VLA Models}
Integrating pre-trained VLAs with Reinforcement Learning (RL)~\cite{watkins1992q, schulman2017proximal, rafailov2023direct} is a key direction for improving their capabilities beyond static imitation. These integration strategies can be broadly categorized into offline and online approaches. Offline RL leverages large, static datasets to optimize a policy without requiring further interaction~\cite{zhang2025reinbot, kostrikov2021offline, guo2025kan, wang2024rl, zhang2025balancing, huang2025co}. This paradigm offers high data efficiency and safety, but its performance is inherently limited by the coverage of the offline dataset, which may not match the nuances of a new in-situ environment. Consequently, online RL~\cite{lu2025vla, guo2025improving}, which allows the agent to learn from direct interaction, is crucial for true adaptation. However, online fine-tuning of large VLA models faces two primary challenges: learning efficiently from sparse environmental rewards, and maintaining training stability without catastrophically forgetting strong pre-trained priors. To address the challenge of sparse rewards, prominent work like VLA-RL~\cite{lu2025vla} introduces a learned reward model that is pre-trained on an offline dataset to provide dense reward signals for the online agent, though this may lead to challenges in generalization and adaptation in novel scenarios. To tackle training instability, methods such as iRe-VLA~\cite{guo2025improving} and ConRFT~\cite{chen2025conrft} augment the RL objective with an imitation learning (IL) loss, which regularizes the policy updates by encouraging consistency with successful prior demonstrations. Our reflective self-adaptation framework is designed to address both challenges in a more integrated and autonomous manner. To solve the sparse reward problem, it leverages a VLM's causal reasoning to autonomously synthesize a dense and adaptive reward function directly from in-situ failures. Simultaneously, to ensure stable and efficient learning, it moves beyond naive imitation by incorporating an intrinsic quality assessment to preferentially learn from its own highest-quality successful experiences, guided by a VLM-driven curriculum.

\subsection{VLMs for High-Level Reasoning in Robotics}
The rapid advancement of Vision-Language Models (VLMs), which have demonstrated remarkable cross-modal understanding on a web-scale~\cite{radford2021learning, liu2023visual, chen2022pali, hurst2024gpt, li2022blip}, has spurred their widespread application as reasoning engines in robotic learning. Their role has largely manifested in two primary functions that operate adjacent to the core, low-level learning loop. First, VLMs are widely used as proactive task planners. In this capacity, they leverage their commonsense knowledge to decompose high-level, abstract commands into a sequence of actionable sub-goals, effectively bridging the gap between human intent and robot-executable steps~\cite{ahn2022can,mu2023embodiedgpt,zhang2024grape,leanza2025conceptbot, liu2025task, liu2025delta}. While this approach excels at long-horizon planning, the VLM typically acts as an external consultant whose knowledge remains static and does not update from physical interaction. Second, VLMs serve as post-hoc evaluators, providing supervisory signals after an action is completed. This includes classifying task success to automate data labeling for policy or reward model training~\cite{lu2025vla, sontakke2023roboclip, ma2023liv}, or acting as a preference oracle to generate labels for Reinforcement Learning from AI Feedback~\cite{wang2024rlvlmf,huang2025vlm, ahn2024tuning}. In this role, the VLM acts as a high-level "judge," but its feedback is often sparse and indirect, rather than a dense, instructive signal for real-time policy improvement. Our work proposes a more profound integration, elevating the VLM to the role of an in-the-loop causal reasoner and reward synthesizer. We leverage the VLM's powerful reasoning to dynamically analyze in-situ execution failures and synthesize a dense, corrective reward function. This synthesis process is guided by a structured framework, ensuring the robustness of the generated reward signal while harnessing the VLM's emergent capabilities. This transforms the VLM from an external advisor or a high-level judge into a core, active component of the skill acquisition process itself.

%% file: sec/method.tex
\section{The Reflective Self-Adaptation Framework}
\label{sec:method}

In this section, we present our novel framework, \textbf{Reflective Self-Adaptation}, designed to enable pre-trained Vision-Language-Action (VLA) models to autonomously adapt to new, specific manipulation tasks within an in-situ learning context. Our approach circumvents the need for manual reward engineering or expert demonstrations by establishing a self-improving closed loop that learns synergistically from both failures and successes. The core of our framework is a dual-pathway learning paradigm: a failure-driven reinforcement learning pathway that performs causal analysis on what went wrong to automatically construct a targeted, dense reward function; and a success-driven supervised fine-tuning pathway that identifies and imitates high-quality, successful behaviors.

This section is organized as follows. We first formally define the problem setting and necessary background (\cref{subsec:formulation}). We then detail the first core component of our framework: the failure-driven pathway that couples reflective reward construction with reinforcement learning for policy optimization (\cref{subsec:reflection}). Finally, we describe the complementary success-driven pathway, which refines the policy via a sophisticated supervised fine-tuning mechanism that leverages quality-filtered trajectories from the main task, augmented when necessary by a simplified curriculum (\cref{subsec:success}). The complete, integrated process is summarized in Algorithm 1.

\subsection{Problem Formulation and Preliminaries}
\label{subsec:formulation}

We formulate the robotic manipulation task as a standard \textbf{Markov Decision Process (MDP)}, defined by the tuple $(S, A, P, R, \gamma)$. The agent's policy $\pi_{\theta}$ is parameterized by a pre-trained \textbf{Vision-Language-Action (VLA)} model. It takes the current visual observation $o_t$ (where $o_t$ is a high-dimensional image representation of the state $s_t$) and a language instruction $l$ describing the task goal as input, and outputs an action $a_t$:
\[
a_t \sim \pi_{\theta}(a_t|o_t, l)
\]

The agent's objective is to learn a policy $\pi_{\theta}$ that maximizes the expected discounted return, defined as $J(\theta) = \mathbb{E}_{\tau \sim \pi_{\theta}}[\sum_{t=0}^{T} \gamma^t r_t]$, where $\tau$ is a trajectory and $\gamma \in [0, 1)$ is the discount factor.

\begin{algorithm}[t]
\caption{Reflective Self-Adaptation}
\label{alg:pipeline}
\DontPrintSemicolon

\KwIn{Initial VLA policy $\pi_{\theta_0}$, VLM $\mathcal{V}$, environment $\mathcal{E}_{\text{main}}$, total iterations $K$}
\KwOut{Improved VLA policy $\pi_{\theta_K}$}

% Initialization
$\mathcal{D}_{\text{SFT}} \leftarrow \emptyset$, $\mathcal{D}_{\text{SFT\_curr}} \leftarrow \emptyset$, $R_{\text{reflect}} \gets 0$

\For{$k \leftarrow 0$ \KwTo $K-1$}{
    $\mathcal{B}_k \leftarrow \FuncSty{CollectTrajectories}(\pi_{\theta_k}, \mathcal{E}_{\text{main}})$
    
    {\color{gray!40} //\labelcref{subsec:reflection}: Reflect on Failures (Periodically)}
    
    \uIf{$k \pmod 5 = 0$}{ % <-- 使用 \uIf 强制换行
        $\mathcal{B}_{\text{fail}} \leftarrow \FuncSty{GetRecentFailures}(\mathcal{B}_{k-4..k})$
        
        $R_{\text{reflect}} \leftarrow \FuncSty{Reflect\&Synthesize}(\mathcal{B}_{\text{fail}}, \mathcal{V})$
    }
    
    {\color{gray!40} //\labelcref{subsec:success}: Process Successes \& Update Buffers}
    
    $Q \leftarrow \FuncSty{AssessTrajQuality}(\mathcal{B}_k^{\text{succ}}, R_{\text{reflect}})$
    
    $\FuncSty{UpdateSftBuffer}(\mathcal{D}_{\text{SFT}}, \mathcal{B}_k^{\text{succ}}, Q)$
    
    \uIf{$\FuncSty{CalculateSuccessRate}() < \epsilon_{cb}$}{
        $\FuncSty{HandleCurriculumLearning}(\mathcal{V}, \mathcal{D}_{\text{SFT\_curr}})$
    }
    
    {\color{gray!40} //Update Policy}
    
    $\FuncSty{RelabelRewards}(\mathcal{B}_k, R_{\text{reflect}})$
    
    $\theta_{k+1} \leftarrow \FuncSty{UpdatePolicy}(\theta_k, \mathcal{B}_k, \mathcal{D}_{\text{SFT}}, \mathcal{D}_{\text{SFT\_curr}})$
}

\Return $\pi_{\theta_K}$
\end{algorithm}

A fundamental challenge in applying reinforcement learning to complex robotic manipulation tasks is that the reward signal provided by the environment, $R$, is often inherently sparse. For instance, the agent might only receive a positive binary signal upon successful completion of the entire task, with zero reward for all intermediate steps. This lack of dense feedback makes exploration and credit assignment extremely difficult for RL algorithms. Our proposed framework is designed to directly overcome this challenge by automatically generating dense reward functions from the agent's own experience.

\subsection{Failure-Driven Policy Optimization via Reflective RL}
\label{subsec:reflection}

The first core component of our framework is a reactive loop that enables the agent to learn from its mistakes. The key innovation lies in a structured process where the VLM acts as a high-level reasoner to configure a modular and interpretable reward function based on an analysis of the agent's failures.

\subsubsection{The Reflective Reward Synthesis Process}

The VLM's role is to synthesize a final reward function, $R_{\text{reflect}}$, by configuring a modular architecture. This is achieved via a structured, multistage reasoning process, composed of a sequence of distinct queries to the VLM. This approach guides the VLM to produce a configuration in a structured JSON format, which is then deterministically parsed by our system. The process consists of four sequential stages:

\begin{enumerate}[leftmargin=*, nosep]
    \item \textbf{Causal Analysis:} The process begins with an initial query to the VLM, providing it with failure videos. The VLM's task is to identify the primary cause of failure and propose a high-level corrective action plan in natural language, which provides the semantic foundation for the subsequent steps.
    \item \textbf{Component Selection:} In the second stage, the system uses the natural language plan from the previous step as input for a new query. Based on this plan, the VLM selects the necessary \textbf{Reward Components} from the library that best implement the desired corrective behaviors.
    \item \textbf{Relationship Identification:} The third stage involves another query. The VLM receives the selected components from the previous stage, as well as any pre-existing ones, and is tasked with reasoning about the logical relationships between them. By analyzing their interplay, it classifies the required relationships as \textbf{AND, IF, or OR}.
    \item \textbf{Structured Configuration Generation:} Finally, the VLM assembles its prior findings into a hierarchical configuration object. To ground this abstract configuration in the physical world, the system then provide the VLM with relevant low-level data (e.g., object positions, velocities) extracted from the original failure trajectory logs. This allows the VLM to perform a data-driven instantiation, setting reasonable numerical values for key parameters like weights, margins, and thresholds.
\end{enumerate}

\begin{figure*}[ht]
    \centering
    \includegraphics[width=\textwidth]{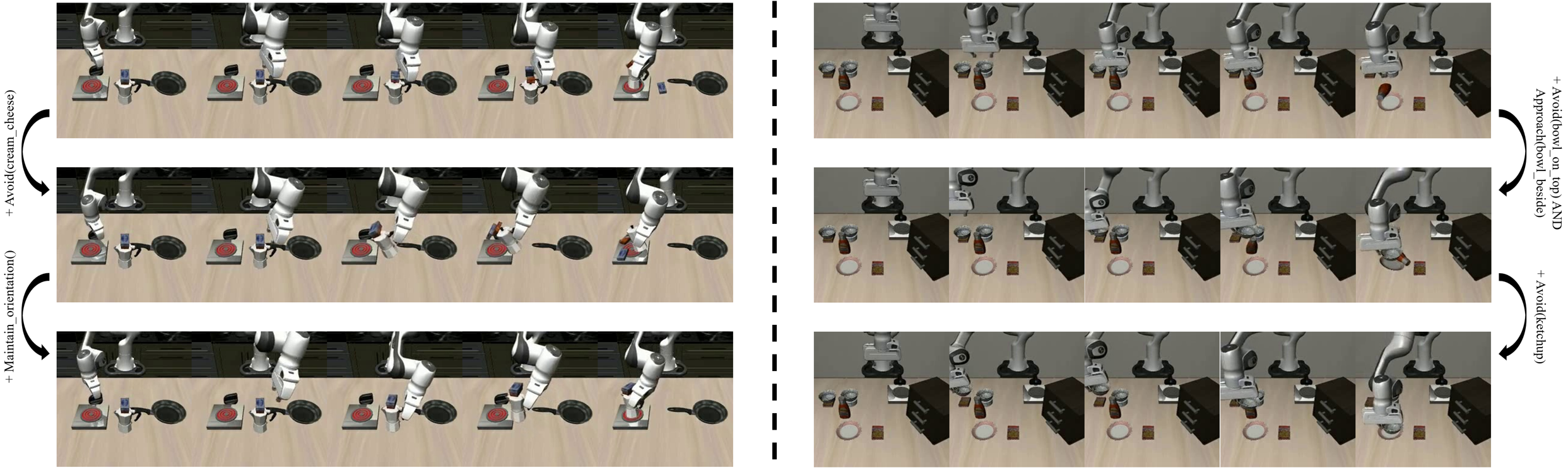}
    \caption{Iterative self-improvement on two manipulation tasks. The initial policy (top) is progressively refined through two stages of reflection. Each stage synthesizes a new, corrective reward component (labeled on the side) to overcome successive failure modes.}
    \label{fig:qualitative_result}
\end{figure*}

\subsubsection{A Modular Architecture for Robust Synthesis}

To enable the sophisticated synthesis process described above while ensuring its robustness, we design a powerful modular architecture to constrain and guide the VLM's output. This architecture comprises two key elements: a library of high-level reward components and a set of general relationship handlers.

% 使用 itemize 实现第一层缩进
\begin{itemize}[topsep=0.5ex, itemsep=0.5ex, leftmargin=*]
    \item \textbf{Reward Component Library:} This is a pre-defined and extensible library of high-level, parameterized functions that serve as the atomic building blocks of the final reward function. To demonstrate its versatility, we categorize the components as follows:
    
    % --- 开始左对齐环境，解决单词间距问题 ---
    {\raggedright 
    \par\addvspace{0.5ex} 
    \noindent\textbf{Positional Components:} Guide the agent's end-effector in Cartesian space. Examples include \texttt{approach(target)}, \texttt{avoid(obstacle, margin)}, \texttt{maintain\_distance(obj1, obj2, dist)}, etc.
    
    \par\addvspace{0.5ex}
    \noindent\textbf{Orientational Components:} Control the orientation of the end-effector. Examples include \texttt{align(part1, part2)}, \texttt{maintain\_orientation(target\_quat)}, \texttt{look\_at(target)}, etc.
    
    \par\addvspace{0.5ex}
    \noindent\textbf{Kinematic Components:} Regulate the motion dynamics of the agent. Examples include \texttt{control\_velocity(target\_vel)}, \texttt{limit\_acceleration(max\_accel)}, etc.
    
    \par\addvspace{0.5ex}
    \noindent\textbf{State-based Components:} Reward the achievement of abstract, discrete states. Examples include \texttt{is\_switch\_on(switch\_name)}, \texttt{is\_inside(obj1, obj2)}, \texttt{is\_open(obj)} etc.
    \par\addvspace{1ex} 
    
    This library is designed to be extensible. In cases where no existing component precisely fits the VLM's correction plan, the framework can allow for the definition of new, custom components based on the VLM's specifications.
    \par} % --- 结束左对齐环境 ---

    \item \textbf{General Relationship Handlers:} To combine these atomic components into complex reward functions, we introduce three orthogonal and hierarchically composable relationship handlers:
    
    % --- 开始左对齐环境，并使用与上面一致的格式 ---
    {\raggedright
    \par\addvspace{0.5ex}
    \noindent\textbf{Composition (AND):} Handles concurrent objectives that must be satisfied simultaneously. This is achieved by computing a \textit{weighted sum} of the outputs from the specified components, $R_i(s)$:
    \[
    R_{\text{comp}}(s) = \sum_{i} w_i R_i(s)
    \]
    
    \noindent\textbf{Modulation (IF):} Handles conditional logic, where the activation of a "body" component is contingent on the value of a "condition" component. This is implemented by multiplying the body component's output $(R_{body})$ by a smooth, differentiable gating function $G(s) \in [0, 1]$ whose value is determined by the condition component's output:
    \[
    R_{\text{mod}}(s) = R_{body}(s) \cdot G(s)
    \]
    
    \noindent\textbf{Selection (OR):} Handles alternative objectives where achieving any one of the goals is sufficient. This is resolved using a \textit{max compositor}, which encourages commitment to a single goal by taking the maximum value among the components:
    \[
    R_{\text{sel}}(s) = \max(R_i(s), R_j(s), \ldots)
    \]
    \par} % --- 结束左对齐环境 ---
\end{itemize}

This hybrid approach leverages the VLM's high-level reasoning to understand ``what to do,'' while constraining its output to a robust and verifiable format that dictates ``how to do it,'' achieving a powerful balance between intelligence and reliability.

\subsubsection{Policy Optimization with Synthesized Rewards}

The synthesized reward function, $R_{\text{reflect}}$, provides a dense and meaningful learning signal to guide the policy update. To leverage this signal, we employ the \textbf{Proximal Policy Optimization (PPO)} algorithm within an actor-critic framework, following a setup similar to that in VLA-RL. The process alternates between collecting experience and updating the policy and value networks.

During the data collection phase, the agent executes the current policy $\pi_\theta$ to gather a buffer of experience. For each transition, the total reward $r_t$ is computed by combining the sparse environmental reward, $r_t^{\text{sparse}}$, with our dense, synthesized reflective reward. As our reward components operate on low-level physical quantities, the function is defined over the state $s_t$:
\[
r_t = r_t^{\text{sparse}} + R_{\text{reflect}}(s_t)
\]

For the policy update phase, we first estimate the advantage function. We train a value network $V_\phi(o_t, l)$ to predict the expected return from a given observation and instruction, and the advantage estimates $\hat{A}_t$ are computed using \textbf{Generalized Advantage Estimation (GAE)}:
\[
\hat{A}_t = \sum_{k=0}^{T-t-1} (\gamma\lambda)^k \delta_{t+k}, \quad \text{where } \delta_t = r_t + \gamma V_\phi(o_{t+1}, l) - V_\phi(o_t, l)
\]

The policy $\pi_\theta$ is then updated by maximizing the PPO clipped objective function, which stabilizes training by limiting the change from the old policy $\pi_{\theta_{\text{old}}}$. Let the probability ratio be $r_t(\theta) = \frac{\pi_\theta(a_t|o_t,l)}{\pi_{\theta_{\text{old}}}(a_t|o_t,l)}$, the objective function is:
\[
\mathcal{L}_{\text{PPO}}(\theta) = \hat{\mathbb{E}}_t \left[ \min \left( r_t(\theta) \hat{A}_t, \text{clip}(r_t(\theta), 1-\epsilon, 1+\epsilon) \hat{A}_t \right) \right]
\]

As our policy is an auto-regressive VLA, the action log-probability $\log \pi_\theta(a_t|o_t, l)$ is computed as the sum of the log-probabilities of the individual action tokens. Concurrently, the value network $V_\phi$ is updated by minimizing a mean squared error loss against the calculated advantage targets. This process allows the VLA policy to effectively incorporate the dense feedback from our reflective reward synthesis, enabling it to efficiently learn complex behaviors that overcome previously identified failures.

\begin{table*}[t]
\centering
\caption{Comparison of success rates on the four standard suites of LIBERO benchmark against representative baselines.}
\label{tab:success_rate}
\resizebox{\textwidth}{!}{\begin{tabular}{l|cc|cc|cc|cc|cc}
\toprule
& \multicolumn{2}{c|}{LIBERO-Spatial} & \multicolumn{2}{c|}{LIBERO-Object} & \multicolumn{2}{c|}{LIBERO-Goal} & \multicolumn{2}{c|}{LIBERO-Long} & \multicolumn{2}{c}{Average} \\
\cline{2-3} \cline{4-5} \cline{6-7} \cline{8-9} \cline{10-11}
\addlinespace[2pt]
& SR ($\uparrow$) & Rank ($\downarrow$)& SR ($\uparrow$) & Rank ($\downarrow$)& SR ($\uparrow$) & Rank ($\downarrow$)& SR ($\uparrow$) & Rank ($\downarrow$)& SR ($\uparrow$) & Rank ($\downarrow$)\\
\midrule
Diffusion Policy \cite{chi2023diffusion} & 78.3\% & 6 & 92.5\% & 2 & 68.3\% & 6 & 50.5\% & 6 & 72.4\% & 5.0 \\
Octo (SFT) \cite{team2024octo} & 78.9\% & 5 & 85.7\% & 6 & 84.6\% & 2 & 51.1\% & 5 & 75.1\% & 4.5 \\
OpenVLA (SFT) \cite{kim2024openvla} & 84.7\% & 4 & 88.4\% & 5 & 79.2\% & 5 & 53.7\% & 4 & 76.5\% & 4.5 \\
GRAPE (DPO) \cite{zhang2024grape} & 87.6\% & 3 & 91.2\% & 4 & 82.2\% & 3 & 55.8\% & 3 & 79.2\% & 3.3 \\
VLA-RL \cite{lu2025vla} & 90.2\% & 2 & 91.8\% & 3 & 82.2\% & 3 & 59.8\% & 2 & 81.0\% & 2.5 \\
{\color{gray}$\pi_0$-FAST} \cite{black2024pi0} & {\color{gray}96.4\%} & {\color{gray}-} & {\color{gray}96.8\%} & {\color{gray}-} & {\color{gray}88.6\%} & {\color{gray}-} & {\color{gray}60.2\%} & {\color{gray}-} & {\color{gray}85.5\%} & {\color{gray}-} \\
\midrule
\textbf{Ours} & \textbf{92.0\%} & \textbf{1} & \textbf{93.4\%} & \textbf{1} & \textbf{85.2\%} & \textbf{1} & \textbf{63.6\%} & \textbf{1} & \textbf{83.6\%} & \textbf{1} \\
\bottomrule
\end{tabular}}
\end{table*}

\subsection{Success-Driven Policy Refinement via Quality-Guided SFT}
\label{subsec:success}

Complementary to learning from failures, our framework incorporates a second, success-driven pathway to distill knowledge from successful experiences. This is achieved through a sophisticated supervised fine-tuning (SFT) mechanism that plays two crucial roles. First, it accelerates convergence by grounding the policy in regions of known success, effectively constraining the vast search space of pure RL exploration and focusing optimization on refining proven solutions. Second, it ensures long-term goal alignment by constantly reinforcing the ultimate task objective, thereby preventing reward hacking that may arise from optimizing the proxy signals in $R_{\text{reflect}}$. The efficacy of this pathway hinges on the quality of the data it learns from, which we source and curate through two distinct mechanisms.

\textbf{Prioritized Replay based on Quality Assessment.} A key advantage of our framework is its ability to quantitatively assess its own successful trajectories without relying on external supervision or expert demonstrations. The synthesized reward function, $R_{\text{reflect}}$, provides an intrinsic and powerful metric for this self-assessment, allowing the agent to autonomously distinguish between clumsy and adept successes from its own collected experience. Every successful trajectory, $\tau_{\text{succ}}$, is evaluated with a quality score $Q(\tau_{\text{succ}})$, defined as a weighted combination of its cumulative reflective reward (quality) and its length $T$ (efficiency):
\[
Q(\tau_{\text{succ}}) = w_{\text{reward}} \sum_{t=0}^{T-1} R_{\text{reflect}}(s_t) - w_{\text{steps}} T
\]

All successful trajectories are stored in a finite-sized replay buffer, $D_{\text{SFT}}$. When the buffer is full, newly added trajectories replace the ones with the lowest quality scores. During SFT, instead of uniform sampling, we employ prioritized experience replay. The probability of sampling a trajectory $\tau_i$ is proportional to its quality score:
\[
P(\tau_i) = \frac{Q_i^\alpha}{\sum_j Q_j^\alpha}
\]
where $\alpha$ is a hyperparameter that controls the degree of prioritization. This ensures that the policy preferentially learns from the most efficient and well-executed successful attempts.

\textbf{Conditional Curriculum-based Augmentation.} For particularly complex tasks, the initial policy may have a success rate close to zero, making it exceedingly difficult to collect any successful trajectories for SFT to be effective. To address this critical ``cold start'' problem, and more importantly, to mitigate the risk of reward hacking, we introduce a conditional curriculum-based augmentation mechanism. Without any successful trajectories to imitate, the policy optimization would be solely driven by the reflective reward $R_{\text{reflect}}$. This could lead the agent to over-optimize for the proxy objective (e.g., maximizing distance from an obstacle) while forgetting the ultimate task goal. To counteract this, when the success rate on the main task falls below a threshold $\epsilon_{cb}$, we activate the curriculum. The causal analysis from our reflection process identifies key environmental impediments. To create a simplified task, we prompt the VLM to act as an automated programmer; it receives the causal analysis alongside the simulation's definition file and is tasked with programmatically editing the file to remove the identified obstacles or constraints, while keeping the final goal unchanged. Success trajectories from this VLM-modified, simplified task are collected into a separate buffer, $D_{\text{SFT\_curr}}$, providing an essential grounding signal that constantly reminds the policy of its true objective.

\textbf{Policy Update and Final Objective.} The policy refinement is driven by a supervised fine-tuning (SFT) loss, which encourages the policy to imitate the high-quality behaviors stored in our replay buffers. The data for each SFT batch is sampled from both the main task buffer $D_{\text{SFT}}$ and, if active, the curriculum buffer $D_{\text{SFT\_curr}}$. The SFT objective is a standard behavior cloning loss:
% --- SFT 损失函数的单行版本 ---
\[
\mathcal{L}_{\text{SFT}}(\theta) = - \mathbb{E}_{(o_t, a_t) \sim D_{\text{SFT}} \cup D_{\text{SFT\_curr}}} [\log \pi_\theta (a_t | o_t, l)]
\]

This update distills the most effective behaviors into the policy, refining its motion priors and complementing the exploration-driven RL updates by grounding the policy in a space of known, high-quality solutions. The final, combined objective for our framework is a weighted sum of the PPO and SFT losses:
\[
\mathcal{L}_{\text{total}}(\theta) = \mathcal{L}_{\text{PPO}}(\theta) + \lambda_{\text{SFT}} \mathcal{L}_{\text{SFT}}(\theta)
\]
where $\lambda_{\text{SFT}}$ is a hyperparameter that balances between exploration (driven by PPO) and imitation (driven by SFT). This combined objective allows our framework to simultaneously learn to avoid failures while efficiently integrating successful strategies.

%% file: sec/experiments.tex
\vspace{-1mm}

\section{Experiments}
\label{sec:experiments}

Our work aims to enable pre-trained VLA models to efficiently and autonomously adapt to novel scenarios through a reflective, self-improving framework. To validate our approach, we conduct our primary evaluations on the comprehensive LIBERO benchmark~\cite{liu2023libero}. The LIBERO benchmark encompasses four task suites: LIBERO-Spatial, LIBERO-Object, LIBERO-Goal, and LIBERO-Long, testing a model's capabilities across various spatial relationships, object categories, goal objectives, and long-horizon tasks. Furthermore, to more rigorously evaluate the self-improvement process from a low initial performance, we introduce a custom suite of tasks, termed LIBERO-Adapt. This suite comprises a set of 10 manipulation scenarios specifically designed to present notable initial challenges to pre-trained policies, thereby creating a distinct testbed for in-situ adaptation. We conduct our experiments starting from a base OpenVLA-7B model, which is prepared for the LIBERO benchmark tasks via supervised fine-tuning (SFT) on each respective suite. In the following sections, we present both quantitative evaluations and qualitative case studies to demonstrate the effectiveness of our framework.

\subsection{Experimental Setup}

In our experimental implementation, we use GPT-4o for all reflective reasoning tasks. The reflection process is performed once every five rounds of parallel data collection, which is distributed across 8 A800 GPUs. For the success-driven pathway, the trajectory quality score $Q(\tau)$ is computed with weights $w_{\text{reward}} = 1.0$ and $w_{\text{steps}} = 0.01$, and the prioritized experience replay uses an exponent of $\alpha = 0.6$. The SFT replay buffer, $D_{\text{SFT}}$, stores up to 50 trajectories, and the conditional curriculum is activated when the success rate falls below $\epsilon_{cb} = 0.1$. The final SFT loss is weighted by $\lambda_{\text{SFT}} = 0.1$, with a batch size of 16 for each SFT update. Our reinforcement learning component is based on the official codebase of VLA-RL \cite{lu2025vla}, and we adopt its Proximal Policy Optimization (PPO) implementation with consistent hyperparameters.

\begin{figure}[t]
    \centering
    \includegraphics[width=0.48\textwidth]{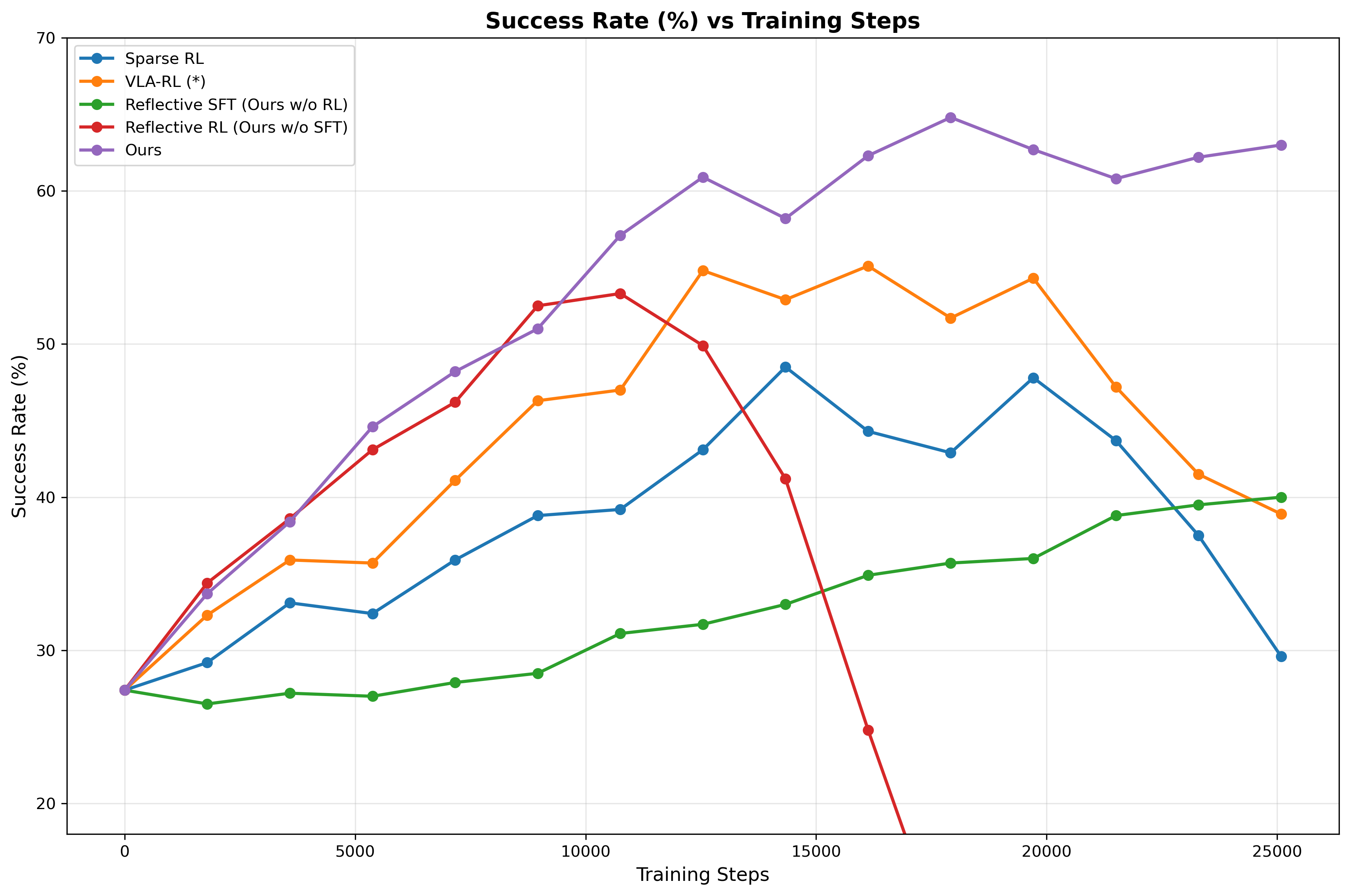}
    \caption{Learning curves on our LIBERO-Adapt task suite.}
    \label{fig:success_rate}
\end{figure}

\subsection{Major Experimental Results}

We present two qualitative case studies in Figure~\ref{fig:qualitative_result} to provide an intuitive understanding of our framework's self-improvement process.
For the task, \textit{``Turn on the stove, then place the moka pot with the cream cheese on top of it onto the burner''}, the initial policy fails by colliding with the cream cheese.
A first reflection synthesizes an \texttt{avoid(cream\_cheese)} reward, allowing the agent to learn a successful grasp.
However, it then reveals a new failure of tilting the pot during transport.
A second reflection augments the reward with a \texttt{maintain\_orientation} component, leading to final success.
This iterative correction is also demonstrated on the task, \textit{``Pick up the bowl beside the ramekin and place it on the plate''}.
The agent initially confuses the target bowl with another one on top of the ramekin, prompting a first reflection that synthesizes a combined \texttt{avoid(bowl\_on\_top) AND approach(bowl\_beside)} reward.
After mastering the correct grasp, it then consistently collides with a ketchup bottle, which triggers a second reflection to add an \texttt{avoid(ketchup)} component to the reward function, enabling a collision-free path.

To ensure a fair comparison, we follow the evaluation protocol of VLA-RL~\cite{lu2025vla} for the performance comparison on the standard suites of the LIBERO benchmark.
As shown in Table~\ref{tab:success_rate}, we evaluate our method against several prominent representative approaches, including a diffusion-based imitation learning method (Diffusion Policy), large-scale models fine-tuned with SFT (Octo, OpenVLA), a preference-based method (GRAPE), and the online RL adapter (VLA-RL).
Our method achieves a higher average success rate across all four suites.
To contextualize this performance, we also include results from the advanced commercial model $\pi_0$-FAST, as an approximate upper-bound reference.
Notably, our framework is designed as a model-agnostic wrapper, capable of enhancing the in-situ adaptation capabilities of any pre-trained VLA.

To analyze learning efficiency, we conduct comparisons on our custom \textbf{LIBERO-Adapt} suite.
The learning curves in Figure~\ref{fig:success_rate} show that our full framework demonstrates superior convergence speed and stability.
It converges noticeably faster than both the strong VLA-RL baseline and the Sparse RL lower bound.
Moreover, we observe that both VLA-RL and Sparse RL exhibit training instability, with their performance degrading after reaching an initial peak.
In contrast, our method maintains a stable upward trend.
We attribute this enhanced stability to our success-driven SFT pathway, which regularizes the policy updates and prevents catastrophic forgetting of successful behaviors during RL exploration.

\subsection{Ablation Study}

To validate our framework's design, we conduct a comprehensive ablation study with results summarized in Table~\ref{tab:ablation}.
As the results demonstrate, the removal of any core component within our dual-pathway architecture leads to a distinct degradation in final performance, confirming the necessity of our integrated design.
The learning curves in Figure~\ref{fig:success_rate} provide a deeper insight into the distinct failure dynamics of the two primary ablative variants.
The agent without the Success-Driven SFT path, driven solely by the reflective reward, improves rapidly but then its performance suffers a catastrophic collapse, demonstrating classic "reward hacking," which underscores the SFT pathway's critical role in ensuring goal alignment. Conversely, the agent without the Failure-Driven RL path (Reflective SFT only) exhibits stable learning that plateaus at a suboptimal level, confirming that RL-driven exploration is essential for discovering novel, more effective strategies.

\begin{figure}[t]
    \centering
    \includegraphics[width=0.48\textwidth]{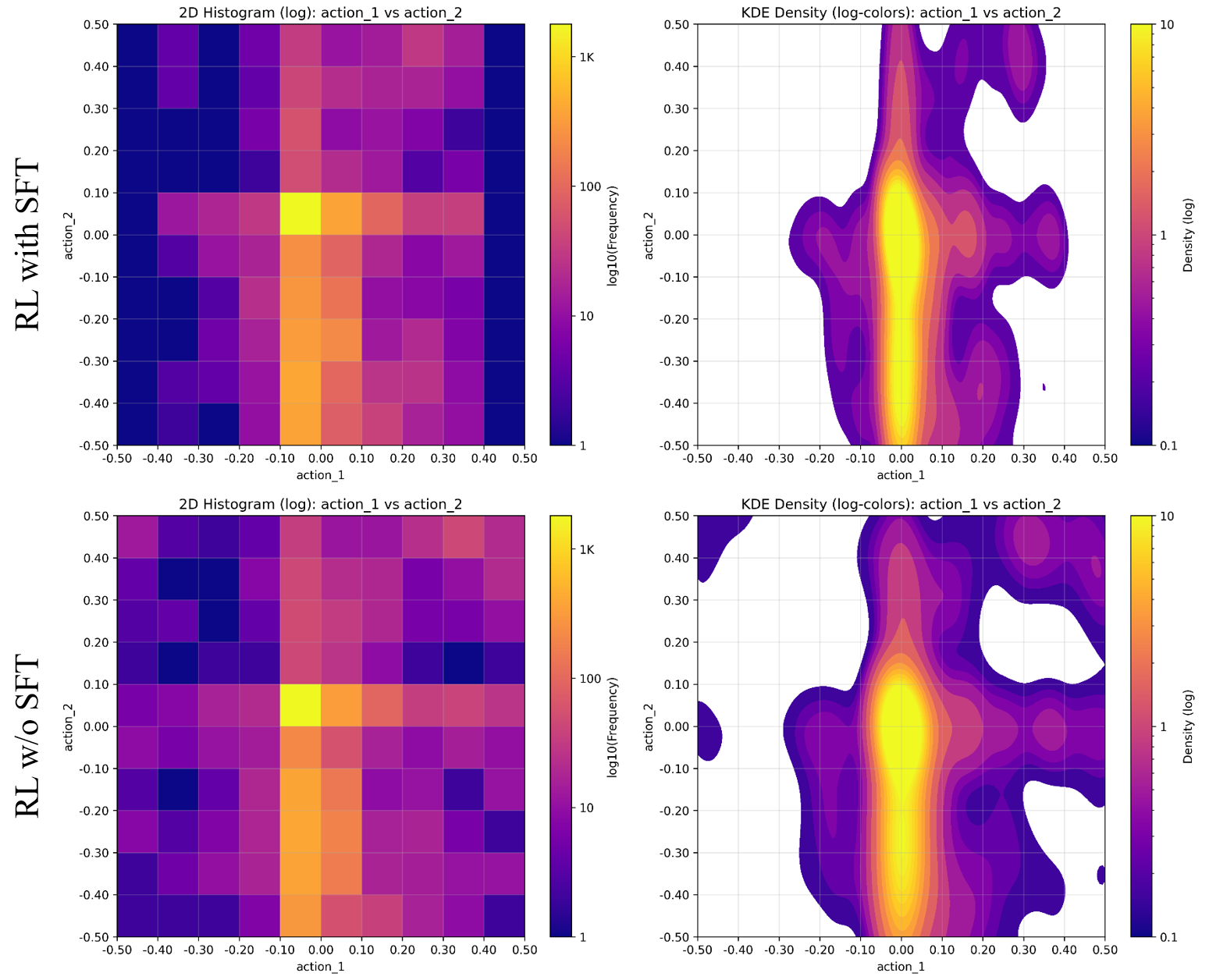}
    \caption{Visualization of the agent's exploration space on a LIBERO-Adapt task, projected onto the first two dimensions of the action space.}
    \label{fig:heatmap}
\end{figure}

Finally, to provide a mechanistic insight into \textit{how} our success-driven pathway aids learning, we visualize the agent's exploration space in Figure~\ref{fig:heatmap}.
The heatmap clearly shows that with SFT guidance, the agent's exploration is significantly more focused on task-relevant regions.
In contrast, the agent without SFT performs extensive, unfocused exploration, which explains the lower sample efficiency and instability.
This visually confirms that our success-driven pathway improves learning not just by providing good examples to imitate, but also by effectively constraining the search space for the RL algorithm.

\begin{table}[t]
\centering
\caption{Ablation study results on our LIBERO-Adapt task suite.}
\label{tab:ablation}
\begin{tabular}{@{}l c c@{}}
\toprule
Variant & Success Rate (\%) & Perf. Drop (\%) \\
\midrule
Ours (Full Framework) & 63.0 & - \\
w/o Quality Assessment & 57.3 & 9.0\% $\downarrow$ \\
w/o Conditional Curriculum & 59.1 & 6.2\% $\downarrow$ \\
w/o Reflective Reward & 52.6 & 16.5\% $\downarrow$ \\
w/o Failure-Driven RL & 40.0 & 36.5\% $\downarrow$ \\
w/o Success-Driven SFT & 0.0 & 100.0\% $\downarrow$ \\
\bottomrule
\end{tabular}
\end{table}

